\documentclass[letterpaper, 10 pt, conference]{ieeeconf}  

\IEEEoverridecommandlockouts                              

\usepackage{graphics} 
\usepackage{epsfig} 
\usepackage{mathptmx} 
\usepackage{times} 
\usepackage{amsmath} 
\usepackage{amssymb}  
\usepackage[T1]{fontenc}
\usepackage{amsfonts}
\usepackage{booktabs}
\usepackage{siunitx}
\usepackage{caption} 
\title{\LARGE \bf
A Markerless Deep Learning-based 6 Degrees of Freedom Pose Estimation for with Mobile Robots using RGB Data
}

\author{Linh K\"astner$^{1}$, Daniel Dimitrov$^{1}$ and Jens Lambrecht$^{1}$
\thanks{$^{1}$Linh K\"astner, Daniel Dimitrov and Jens Lambrecht are with the Chair Industry Grade Networks and Clouds Department, Faculty of Electrical Engineering, and Computer Science,
        Technical University of Berlin, Berlin, Germany
        {\tt\small linhdoan@tu-berlin.de}}%
}

\begin{document}

\maketitle
\thispagestyle{empty}
\pagestyle{empty}

\begin{abstract}
Augmented Reality has been subject to various integration efforts within industries due to its ability to enhance human machine interaction and understanding. 
Neural networks have achieved remarkable results in areas of computer vision, which bear great potential to assist and facilitate an enhanced Augmented Reality experience. However, most neural networks are computationally intensive and demand huge processing power thus, are not suitable for deployment on Augmented Reality devices.
In this work we propose a method to deploy state of the art neural networks for real time 3D object localization on augmented reality devices. As a result, we provide a more automated method of calibrating the AR devices with mobile robotic systems. 
To accelerate the calibration process and enhance user experience, we focus on fast 2D detection approaches which are extracting the 3D pose of the object fast and accurately by using only 2D input. The results are implemented into an Augmented Reality application for intuitive robot control and sensor data visualization.
For the 6D annotation of 2D images, we developed an annotation tool, which is, to our knowledge, the first open source tool to be available. 
We achieve feasible results which are generally applicable to any AR device thus making this work promising for further research in combining high demanding neural networks with \textit{Internet of Things} devices.


\end{abstract}

\section{INTRODUCTION}

The need of an environmental perception and understanding is essential for tasks such as autonomous driving, Augmented Reality (AR) and mobile robotics. 
Mobile robots are one of the main systems to profit from the recent progress in computer vision research. Their popularity within industries has increased due to their flexibility and the variety of use cases they can operate in. Tasks such as provision of components, transportation, commissioning or the work in hazardous environments are increasingly being executed by such robots \cite{smartfactory}, \cite{mobilerob}. However, operation and understanding of mobile robots is still a privilege to experts \cite{mobilerob2} because they are more complex and thus harder to operate and understand. 
On this account, Augmented Reality (AR) has gained much attention in research due to the high potential and ability to enhance efficiency in human robot collaboration and interaction which had been proved by various scientific publications \cite{hashimoto}.  
AR has the potential to aid the user with help of spatial information and the combination with intuitive interaction technology, e.g. gestures or voice commands \cite{arsystems}. Our previous work focused on AR-based enhancements in user understanding for robotics. In \cite{lambrecht} we simplify robot programming with the help of visualizing spatial information and intuitive gesture commands. In  \cite{kaestner} we developed an AR-based application to control the robot with gestures and visualize its navigation data, like robot sensors, path planing information and environment maps. Other work used AR for enhanced visualization of robot data or for multi modal teleoperation \cite{ar1}, \cite{armultimodal}. 
The key aspect to facilitate AR, is the initial calibration between AR device and the components within the environment. For our use case, this means that the AR device has to be spatially aligned with the mobile robot in order to visualize data properly. State of the art approaches are relying on marker detection which proves to be accurate but unhandy. Additionally, markers cannot be deployed everywhere especially not in complex dynamic environments for use cases like autonomous driving. Furthermore, occlusion could impact the performance drastically as was proven in our previous work \cite{kaestner2}. For these reasons, neural networks are increasingly being subject of research efforts.
Recently published work focused on the usage of 3D data like depth sensors for better accuracy. However, within AR applications, 3D data is still too unhandy to use because it require additional preprocessing steps, which slows down the overall application pipeline \cite{kaestner1}. Our previous work explored the possibilities of a 3D sensor integration into an AR head mounted device \cite{kaestner2}. Although we could show the feasibility and the localization was accurate, the processing time was very slow which drops user experience. RGB data on the other hand, dont require the complex preprocessing steps which reduces the computation time and enhance user experience. Furthermore, RGB data is most ubiquitous within AR devices like headsets or smartphones thus, our method can cover a wider range of devices. RGB data also contains color information which is helpful for classification scenarios. On this account, we
explore the possibilities of an integration of neural networks
that can localize the robots 6 Degrees of Freedom (DoF) pose solely based on 2D
data input. We compared two different state of the art neural networks and propose an distributed architecture to integrate those into our AR use case. 
Our results can be adapted for every other AR system which opens many possibilities for further research. 
The main contributions of this work are following:
\begin{itemize}
	\item Proposal of a distributed architecture to deploy neural networks (NNs) for markerless pose estimation on AR devices
	\item Evaluation of different neural network models in terms of feasibility for deployment within AR applications
	\item Development of an open source 6 DoF annotation tool for 2D data
\end{itemize}
The paper is structured as follows. Sec. II will give an overview of related work. Sec. III will present the conceptional design of our approach while sec. IV will describe the the training process and implementation. Sec. V will demonstrate the results with a discussion in Sec. VII. Finally, Sec. VIII will give a conclusion.

\section{RELATED WORK}
Due to the high potential of AR, it has been considered for integration into various industrial use cases.The essential step to facilitate AR is the calibration between the AR device and environmental components e.g. machines or robots. 
\subsection{AR calibration methods}
Manual calibration using external tools or special equipment is the most straightforward method and requires the user to manually align 2D objects to objects in the real environment which position is known to the calibration system. 
 Azuma and Bishop \cite{azuma1994improving} proposed an AR camera pose estimation by manually aligning fiducials like virtual squares to specific locations in the environment.
 Similar work from Oishi et al. \cite{oishi1996methods} require the user to align calibration patterns to different physical objects which positions are known beforehand. However, manual calibration bear disadvantages in a continuously automated environment due to the necessity to have a human specialist to do the calibration. The external tools themselves can be expensive or are case specific.\\ Recent research focus on automated calibration methods rather than manual ones. Most ubiquitous are marker based approaches. Libraries and toolkits like \textit{Aruco} or \textit{Vuforia} make it possible to deploy self created fiducial markers for pose estimation. The marker is detected and tracked using computer vision and image processing. Especially when working with static robots, marker based approaches are widely used due to the simple setup and competitive accuracy. Several work including Aoki et al. \cite{aoki2018self} or \cite{chu2018helping} et al. relied on a marker based calibration between robot and AR device. 
However, Baratoff et al. \cite{baratoff2002interactive} stated, that the instrumentation of the real environment, by placing additional tools like markers, sensors or cameras is a main bottleneck for complex use cases especially in navigation that are working with dynamically changing environments. Furthermore, deploying additional instruments require more time to setup and are not able to react to changes within the environment. Especially for industrial scenarios, this is the main concern \cite{baratoff2002interactive}. Another factor that drops performance in dynamically changing environments is occlusion of markers which was noted in our previous work \cite{kaestner2}. \\
On this account, marker-less approaches have became a major focus \cite{yan2017application}. This includes model-based approaches which use foreknown 3D models to localize objects and generate the pose estimation for the AR device. This is done using computer vision and frame processing to automatically match the known 3D models on the object from the incoming frame. Subsequently the pose is estimated. Works by Bleser et al. \cite{bleser2006online} were able to establish a competitive approach using CAD models and achieving high accuracy. 
The main bottleneck is, that these approaches are very resource intensive due to the processing of each frame.
In addition, these methods are highly dependent on the CAD models, which are not always available. Furthermore, the approach becomes less flexible, which is crucial in dynamic environments for tasks such as autonomous navigation of mobile robots. \\
 With the recent advances in machine learning algorithms, especially deep learning, and the higher computational capabilities of computers, networks can be trained to recognize 3D objects. 
An overview of current research is provided by Yan et al. \cite{yan2017application}.   
Recent publications by \cite{qi2019deep}, \cite{shi2019pointrcnn} have focused on learning directly from 3D data. However, 3D data often require preprocessing steps to be reliably deployable within an AR device. Furthermore it do  not contain color chanels which could be important for classification. On that account, we focus on 2D approaches to achieve real time performance.
Alhaija et al. \cite{alhaija2018augmented} worked with neural networks for object detection and segmentation for AR-supported navigation. The research closest to our method is proposed by Garon and Lalonde \cite{garon2017deep} which used convolutional neural networks (CNNs) to detect 3D objects. We extend this approach principle by applying more recent neural networks based on \textit{YOLOv3} and integrate them into an AR robotic application. Therefore we choose two state of the art neural networks for our use case, SSPE and BetaPose, which are described in the following.
\subsubsection{SSPE}
\textit{SSPE} is heavily inspired by \textit{YOLO} . Thus, its a fast method for 3D pose estimation. \textit{YOLO} was modified to deliver 8 corners of a 3D bounding box together with its center instead of 4 points, like in the original implementation. After the 2D projections are estimated, the Perspective-n-Point (PnP) algorithm is used to calculate the rotation and translation of the object. \textit{SSPE} can manage up to 50 FPS and does not need any refinement. The accuracy on the \textit{LINEMOD} dataset was 55.95 percent which is competitive for an approach without refinement steps.
\subsubsection{Betapose}
Figure 4.2 shows the \textit{Betaposes} pipeline. It consists of 2 NNs. The first one is
YOLO which detects where the object is on the image. If \textit{YOLO} detects
the object, it cuts it out and gives the cutout image to the Key Point Detector
(KPD) network which marks the 2D position of the Key Points (KPs) it was trained to localize.
Finally, a PnP algorithm is used to recover the 3D pose according to the 2D-3D
relationship of the KPs. Just like \textit{SSPE}, \textit{Betapose} does not
rely on a time-consuming post-processing procedure. The method
can achieve competitive accuracy without any refinement
after pose prediction. \textit{Betapose} could produce up to 25 FPS.

\subsection{Distributed Architecture}
Combining neural networks with AR require an efficient architecture to ensure proper visualization for high user experience. That is because most of currently available AR devices are computationally restricted thus, an outsourcing of demanding services and functions has to be taken into consideration. 
Distributed systems for AR devices are increasingly being researched by
the community. Ren et al. \cite{ren2019edge} and Schneider et al. \cite{schneider2017augmented} both presented different
edge computing architectures for AR use cases. They both put emphasis
on the latency and power consumption connected with the distributed nature
of the system. However, the most extensive and well documented research was presented by Ha et al. \cite{ha2014towards}, which is a system for wearable cognitive assistance on the network edge.
It consists of 2 hardware devices - the wearable AR-enabled device (client) and the server also called \textit{Cloudlet}. A \textit{Cloudlet} is a computer located on the network edge. It can also be described as a "data center in a box" which is one hop away from the AR-enabled device.
Within that server the user can deploy own services which are also called proxies. These proxies can cover a wide range of application ranging from face recognition, motion classifiers or object detectors.
The client is sending different sensory data that it wants to be evaluated to the server. The implemented proxies will evaluate them and returns assimilated data back to the client. One of the advantages of Gabriel is that it is easily extensible with new and multiple services/proxies. It offers tight latency bounds on computationally heavy calculations. Furthermore, it also takes into consideration the limited battery and computational capacity of
the AR-enabled devices, which is why it is most suitable for our use case.

\section{CONCEPTUAL DESIGN}
In this chapter, we present the conceptual design of our work. The modules and approaches used are based on the research done in the related work section.

\subsection{AR Device Calibration}
To facilitate AR, the AR device has to be aligned with its environment. Our method assists in robotic environment hence the AR device has to be properly aligned with the robot entities.
For our use case, both entities, AR device and robot are fully mobile which means that a continuous localization of the robot has to be ensured.
For the AR application first developed in our previous work \cite{kaestner}, we proposed a two-stage calibration between AR headset and robot. First, the \textit{Hololens} is calibrated with the robot using \textit{Aruco} markers. Second, the spatial anchor functionality of the \textit{Hololens} was used for calibration of the robot with the robot map by placing an spatial anchor at the position of the detected marker. The spatial anchor capability refers to the internal SLAM of the \textit{Hololens} to scan the room and memorize its location within the room. 
The robot map is the general reference map from which all sensor data and information are displayed. Thus, calibration between the AR device and the robot internal map ($T_{AR-Map}$) is the overall goal to visualize data properly within the AR headset. The robot position ($T_{Robot-Map}$) is always known by the map through the robots \textit{SLAM} packages (e.g. \textit{Adaptive Monte Carlo}). Hence, to achieve a transformation between AR device and ROS map, we have to achieve a transformation between AR device and robot ($T_{AR-Robot}$). 
This is shown in equation \ref{eq1}.

\begin{align}
T_{AR-Map} = T_{AR-Robot} \cdot T_{Robot-Map}
\label{eq1}
\end{align}

\begin{figure}[!h]
	\centering
	\includegraphics[width=3.3in]{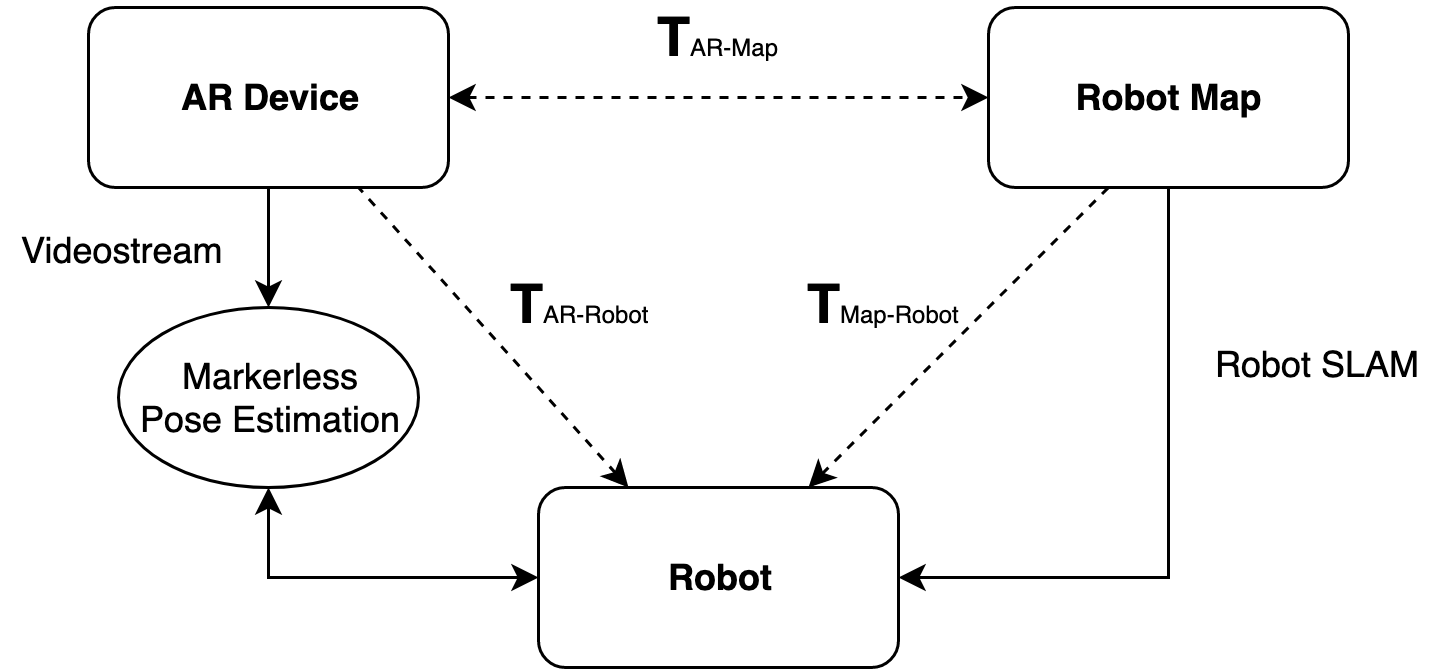}
	\caption{Conception of Coordinate System Alignment}
	\label{conceptcoord}
\end{figure}

With this work, the transformation between AR device and robot $T_{AR-Robot}$ is generated with a neural network-based robot pose estimation thus replacing the marker-based approach from our previous work. Based on that localization, the spatial anchor capability can be set and continuously track the robots position. This concept is illustrated in Fig. \ref{conceptcoord}.

\subsection{Neural Network Selection}
Our AR calibration has to provide a fast localization ob the robot results in order to speed up the spatial alignment process and ensure an enhanced user experience. The neural network should therefore perform fast and accurate results at the same time. Our previous \cite{kaestner2} work explored the performance of 3D data and although we showed promising results and the overall feasibility, the approach required up to 40 seconds of preprocessing time which drops the user experience tremendously. RGB images have the potential to perform faster and are more established with countless state of the art approaches achieving competitive results. Thus, our research will only consider these kind of neural networks that takes RGB data as input. We are inspired by the release of state of the art neural networks working only with 2D information to obtain the 3D position in a fast manner by using single shot object detectors like \textit{YOLO} \cite{redmon2016you}.  Hence, we choose \textit{Betapose} by Zhao et al. \cite{zhao2018estimating}  and \textit{SSPE} by Tekin et al.  \cite{tekin2018real} for our use case as they provide fast and accurate 6 DoF pose estimation using only RGB data. As input, we will take the AR device video-stream for training and evaluation of the neural networks. 

\subsection{Distributed Architecture}
Due to the limited hardware capabilities of the \textit{Microsoft HoloLens}, we
propose a distributed application with two main components - the client (the AR-enabled
device) and the server (a more powerful computer located on the network
edge near the AR-enabled device). The computationally intensive neural network operations have to be outsourced to the server site, whereas visualizations of the output are computed on client site. For our use case we build 2 separate services/proxies using the \textit{Gabriel} framework which are described in the implementation chapter in more detail.

\section{IMPLEMENTATION}
\begin{figure}[h!]
	\centering
	\includegraphics[width=3.4in]{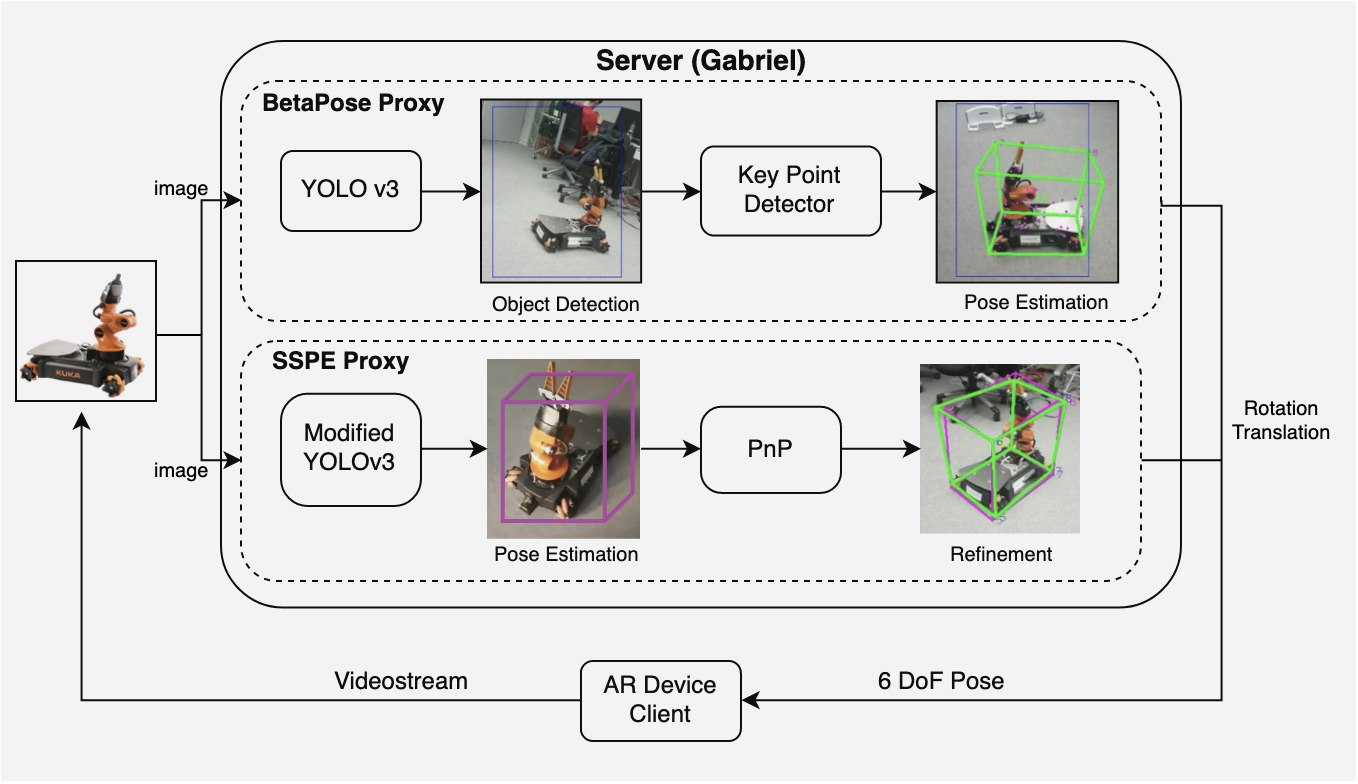}
	\caption{Overall workflow of the application}
	\label{im2}
\end{figure}

We propose a distributed architecture using the open source framework \textit{Gabriel} which allows the user to build servers to run computationally intensive operations like neural networks on it and deploy the results in edge devices. These servers are also called proxies. The overall workflow is depicted in Fig. \ref{im2}. We implemented 2 proxies for the 2 different neural networks.  The implementation for \textit{SSPE} was based on the open source \textit{Github} repository provided by the authors. For our use case, we modified the evaluation code so that it would only take in a single image to evaluate and not a whole
dataset. Additionally to the estimated 3D bounding box projection this proxy
also projects the output of the \textit{SSPE} NN - that is the centroid and the 8 estimated
corners of the bounding box.
The implementation of \textit{Betapose} was more complex compared to \textit{SSPE} since \textit{Betapose} contains two NNs one
feeding it’s result to the next one. The system had to be parallelized for maximum
efficiency by using 3 threads - one for the main program
accepting and responding to HTTP requests, one for the \textit{YOLO} NN and
one for the KPD NN. Additionally to the estimated 3D bounding box projection
our implemented proxy also projects the output of the \textit{YOLO} NN and the KPD NN onto the
\textit{Cloudlet} AR device for enhanced understanding. For the \textit{YOLO} NN that is the 2D bounding box around the
object. For the KPD NN that is the 2D position of each KP.
The two proxies are can be run separately as well as in parallel e.g. for performance comparison.

\subsection*{Hardware Setup}
The hardware setup contains a mobile robot, a powerful Linux computer with Tesla K80 GPU acting as the server and a head mounted AR device - the \textit{Microsoft Hololens}. We are working with a \textit{Kuka Mobile Youbot} running ROS \textit{Hydro} on a \textit{Linux} version 12.04 distribution. As a development environment for the \textit{Hololens}, we use a windows notebook with \textit{Unity3D 2019} installed on which the application was developed. The videostream extraction is done in \textit{CSharp} using the \textit{Mediastream API} provided by \textit{Microsoft} using \textit{Visual Studio 2019} as development environment. 
All entities are connected to the same network and communicate a \textit{WLAN} connection.

\subsection*{Dataset generation}
\begin{figure}[h!]
	\centering
	\includegraphics[width=3.4in]{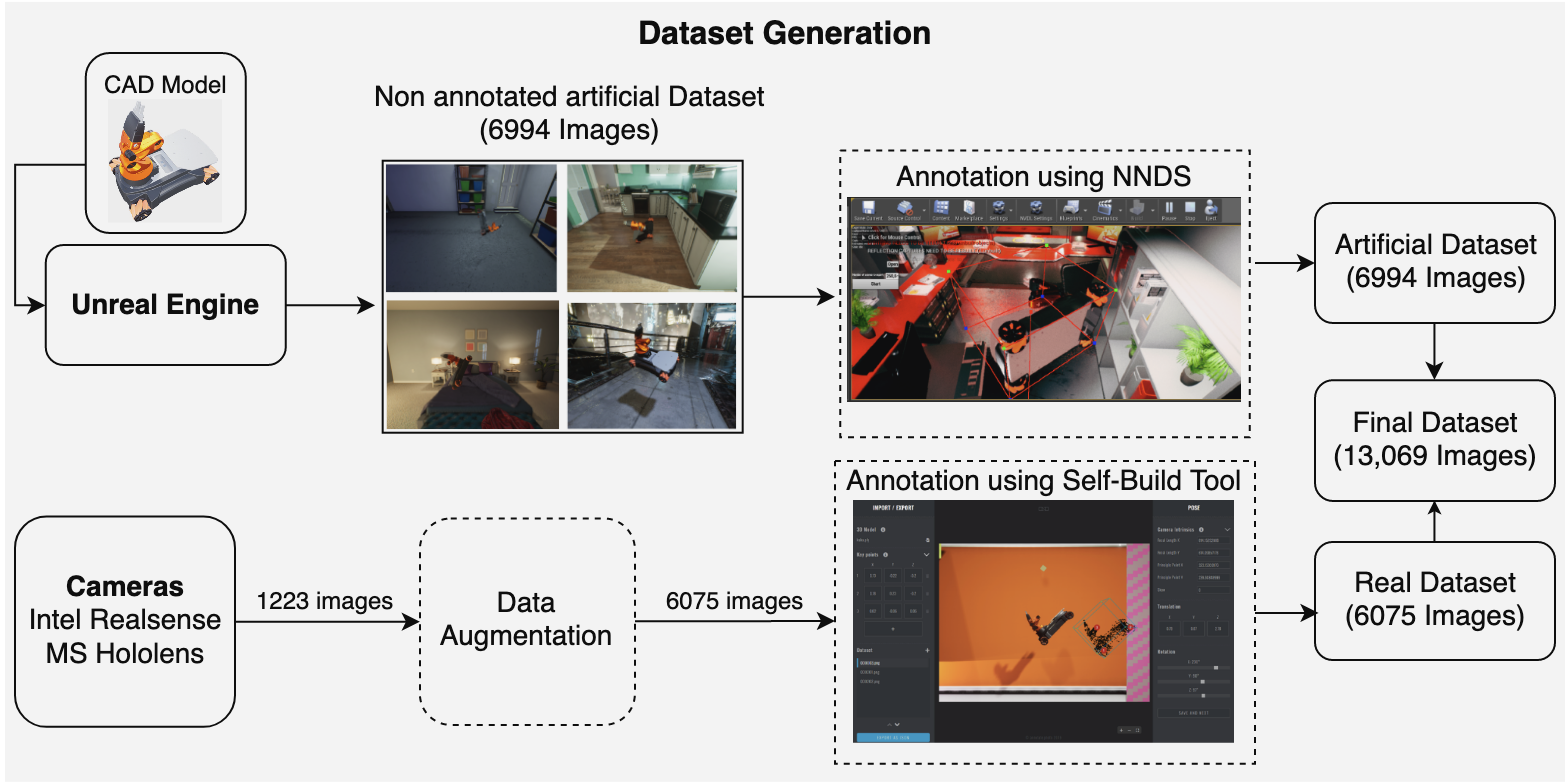}
	\caption{Real and artificial dataset generation pipeline}
	\label{data3}
\end{figure}

For training the neural networks, we created a dataset consisting of real images as well as artificially created ones, to ensure a large dataset for high performance. The process of dataset generation is illustrated in Fig. \ref{data3}. To generate the artificial dataset, we used the game engine \textit{Unreal Engine} and a 3D robot model. We used 18 pre existing environments and placed the robot at different angles as depicted in Fig. \ref{data3} to ensure variety. The annotation of the artificial data was done with a plugin tool from \textit{Unreal Engine 4} called \textit{NDDS} which annotates the 6D pose of the object within the 2D images. 

The real dataset was captured with an \textit{Intel Real Sense} camera as well as with the \textit{Microsoft Hololens} camera stream. 
Additionally, we used data augmentation like flipping, rotation and scaling to enlarge the real dataset. For the annotation of the real data, we developed a new tool \footnote[2]{http://annotate.photo} which required no installation and was as easy to use as possible. This tool can annotate 2D images with the 6 DoF pose using a 3D model.
The user can import a 3D model as well as the 2D dataset into the web application. Afterwards, the user has to define a minimum of 3 points and set the camera intrinsics. Our tool has a build in PnP algorithm to calculate the 6D pose out of these information. Additionally, the pose can be manually be refined. The annotated dataset is exported as a JSON file which is containing the annotations for the whole dataset. The tool was used, to annotate the real data captured with the cameras.
In total, our dataset consists of \textbf{13,069} images with 6994 artificially created images and 6075 real images. Originally, we collected 1223 real images which we augmented using rotation, flipping, scaling and changing the contrast to simulate bad lightning conditions. 
Fig. \ref{model2} depicts the used 3D model of the robot as well as the real robot. It must be noted, that both are not exactly equal and differ in small details. E.g. the arm of the model do not have claws and the overall appearance looks more static compared to the real robot. This is why we considered both, artificially created and real data for our training.

 \begin{figure}[h!]
	\centering
	\includegraphics[width=3.4in]{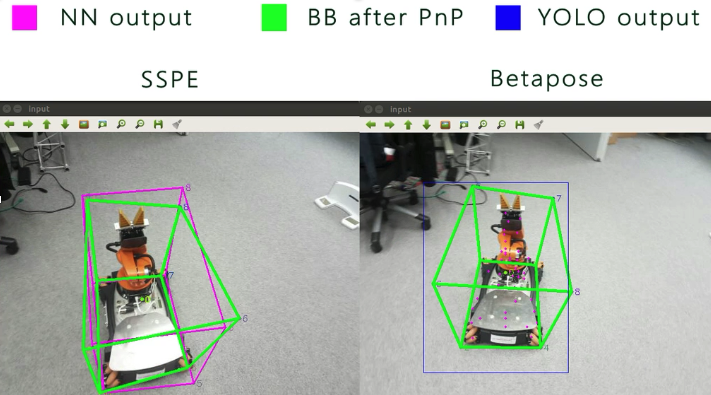}
	\caption{Results from the neural networks - Near. \textit{Left, from SSPE. Right, from Betapose. The user is 0.5 - 1m away from the robot. Both NN can detect the robot and its 6D pose accurately.}}
	\label{res2}
\end{figure}

\subsection{Training process}
For the training a GPU server with \textit{Tesla K80}  with 12 GB GDDR5 memory, 4vCPUs and 15GB RAM was used.
The network was trained for 3700 epochs (where a epoch means one pass through the whole
data set), using batches of size 8 and the Adam optimizer for minimizing the loss function. 
During the second phase, the KPD neural network is retrained using the
weights produced by the first phase and using Pose-guided Proposal Generator
(PGPG) (also called DPG in the \textit{GitHub} repository of \textit{Betapose}) during training.
This PGPG is done to deal with the bias in bounding box detection which would
lead to an error in keypoint localization. It basically means that the dataset is
augmented. Both phases took around 24 hours to complete in total
Training the first phase took around 20 iterations and
around 1 hour to complete. Training the second phase was slower and took
around 135 iterations and around 20 hours to reach peak accuracy.

\section{Evaluation} 
\begin{figure}[]
	\centering
	\includegraphics[width=3.4in]{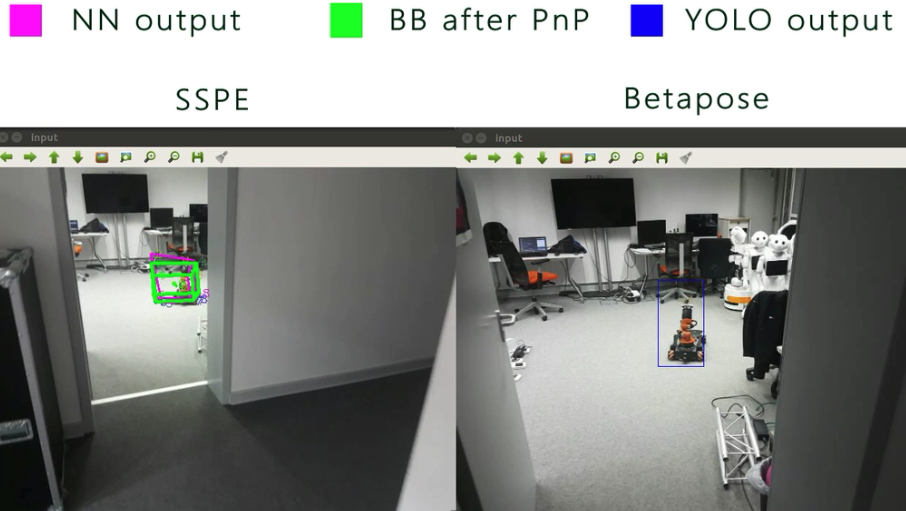}
	\caption{Results from neural networks - Far. \textit{Left, from SSPE. Right, from Beta Pose. In the left picture, the user is 10 meters away from the object. While Betapose can only detect the object, SSPE can still localize its 6D pose.}}
	\label{res1}
\end{figure}

This chapter presents the results of the implemented approaches as well as a discussion of these. 
Fig. \ref{res1} and Fig. \ref{res2} illustrates the results of \textit{Betapose} and \textit{SSPE}. For better understanding, we visualized the output of several stages of the approaches. For Betapose, we visualized the \textit{YOLO} output, the NN output and the bounding box after PnP is done. Furthermore, the KPs are visualized. For \textit{SSPE} we visualized NN output and BB after PnP is done. It can be seen that both methods providing accurate 3D localization of the robot. This is achieved in real time with the video stream of our AR device. Overall \textit{SSPE} is faster achieving 170ms to evaluate the image compared to 500ms for \textit{Betapose}. 
We compared the application with results from our previous work \cite{kaestner2} where we considered a 3D object localization based solely on depth sensor data (point-clouds) as well as our very first approach - the marker detection using \textit{Vuforia} markers with a marker size of 28x28 cm. The comparison is listed in Table \ref{tt}.
All three methods are compared in terms of accuracy, computation time and the maximum distance, where an accuracy of 50 can be still achieved.

\begin{table}[h!]
	\caption{Comparison of implemented approaches}
	\begin{tabular}{lllll}
		\hline
		Metric   & VoteNet(3D) & BetaPose & SSPE   & Marker   \\ \hline
		Accuracy & 95,5        & 93,2         & 92.74      & 91,67    \\ 
		Computing Time    & 6.2s        & 0.5s         & 0.17s      & 0.72s    \\ 
		Distance & $\sim$4m    & $\sim$8m    & $\sim$10 m & $\sim$2m \\ \hline
	\end{tabular}
	
	\label{tt}
	
\end{table}

 The accuracy metric is also called ADD metric in many papers and is an 3D intersection over union. We compared the predictions of the neural networks with our annotated ground truth. The computation time includes all steps - data acquisition, the transmission to the neural network server, the processing and visualization of results on the AR device. The distance was calculated by exploring the space where the accuracy of the approach reach a minimum of 50. The test were done on the \textit{Microsoft Hololens}. All values are average means of 20 different measurements each. It can be observed that the 3D method is the most accurate with 95.5$\%$ in accuracy whereas our recently implemented approach with \textit{SSPE} is by far the fastest approach taking only 0.27s to overall localize the object and display the observings within the AR application. This is taking into consideration the time to send the video-stream from the AR device to the server which takes a total of 0.1s with our proxy implementation. The slowest approach is unsurprisingly the 3D method due to the huge preprocessing steps needed. The extraction of 3D point-clouds from the AR device and processing took 3 seconds while transmitting the data to the server took another 2-3 seconds which results in a long time. Our new neural network-based approach is even faster than the marker-based approach which takes on average 1.32 seconds to detect the robot. This is taking into consideration, that at some positions where the marker is occluded, this approach performed worse which decreased the overall performance even though this approach performed fast at positions where the marker is clearly visible. Another important aspect is that the marker-based approach performed bad when the user was far away and was only working well for a distance up to 2m away from the marker. Whereas the 2D based approach could still provide accurate results with up to 8m as is seen in Fig. \ref{res1}.

\section{CONCLUSION}
We proposed a pipeline to deploy state of the art neural networks for robot pose estimation on capacity limited AR device - the \textit{Microsoft Hololens}. Therefore we made use of a distributed approach with an efficient client server communication. We showed the feasibility of our distributed approach and could replace the classical marker detection approach thus making an AR integration more intuitive and less tedious. We compared two different state of the art neural networks and concluded that \textit{SSPE} is the most suitable network for real time deployment. we obtained remarkable results and compared them with the results of our previous work. Our neural network based object localization outperformed the previously used marker-based approach both in terms of accuracy as well as speed. The 3D based point cloud approach was outperformed in terms of speed. Even though its less accurate then 3D approach, the 3D approach still takes too much time to be considered as competitive for application in industries. The results of this papers are promising for further research in the area of combining neural networks into IoT devices like AR headsets. In future, sensor fusion approaches should be considered to obtain more accurate results.
Furthermore we developed an annotation tool for 6D pose labeling of 2D images which is to our knowledge the first open source tool available online. 

\section{Appendix}
All the code and demo materials are publicly available on \textit{Github} - https://github.com/lee4138/6d-pose-estimation-with-ml-in-ar. The annotation tool is online under the domain http://annotate.photo.
\addtolength{\textheight}{-2cm}   





\bibliographystyle{IEEEtran}

\bibliography{references}

\end{document}